\documentclass[11pt,a4paper]{article}
\usepackage{acl}
\usepackage{times}
\usepackage{latexsym}
\usepackage[TABBOTCAP]{subfigure}
\usepackage[shortlabels]{enumitem}

\usepackage{tikz-dependency}
\usepackage{algorithm}
\usepackage{algpseudocode}
\usepackage{multirow}

\usepackage{color}
\usepackage{helvet}
\usepackage{textcomp}
\usepackage{graphicx}
\graphicspath{ {images/} }
\usepackage{amsmath}
\usepackage{float}
\usepackage{booktabs,amsfonts,dcolumn}
\usepackage{hyperref}
\usepackage{url}

\usepackage[]{collab}
\collabAuthor{yt}{teal}{Yintong Huo}

\def\NM{{\mathcal N}}

\def\0{{\bf 0}}
\def\1{{\bf 1}}

\def\name{{\bf CIKQA}}

\usepackage{xcolor}
\usepackage{soul}
\newcommand{\hlc}[2][yellow]{{%
    \colorlet{foo}{#1}%
    \sethlcolor{foo}\hl{#2}}%
}

\newcommand{\Red}[1]{\textcolor[rgb]{1.00,0.00,0.00}{#1}}
\newcommand{\Blue}[1]{\textcolor[rgb]{0.00,0.00,1.00}{#1}}

\title{CIKQA: Learning Commonsense Inference with a Unified \\ Knowledge-in-the-loop QA Paradigm
}

\author{Hongming Zhang$^{1,2}$, Yintong Huo$^3$, Yanai Elazar$^{4,5}$, Yangqiu Song$^1$, Yoav Goldberg$^{4,5}$, Dan Roth$^2$\\
$^1$HKUST, $^2$UPenn, $^3$CUHK, $^4$AI2, $^5$University of Washington, $^6$Bar Ilan University\\
\texttt{\{hzhangal,yqsong\}@cse.ust.hk}, \texttt{ythuo@cse.cuhk.edu.hk} \\
  \texttt{\{yanaiela,yoav.goldberg\}@gmail.com}, \texttt{danroth@seas.upenn.edu}}
\date{}

\begin{document}
\maketitle
\begin{abstract}

Recently, the community has achieved substantial progress on many commonsense reasoning benchmarks. However, it is still unclear what is learned from the training process: the knowledge, inference capability, or both? 
We argue that due to the large scale of commonsense knowledge, it is infeasible to annotate a large enough training set for each task to cover all commonsense for learning. Thus we should separate the commonsense knowledge acquisition and inference over commonsense knowledge as two separate tasks. 
In this work, we focus on investigating models' commonsense inference capabilities from two perspectives: (1) Whether models can know if the knowledge they have is enough to solve the task; 
(2) Whether models can develop commonsense inference capabilities that generalize across commonsense tasks.
We first align commonsense tasks with relevant knowledge from commonsense knowledge bases and ask humans to annotate whether the knowledge is enough or not.
Then, we convert different commonsense tasks into a unified question answering format to evaluate models' generalization capabilities.
We name the benchmark as Commonsense Inference with Knowledge-in-the-loop Question Answering (\name).

\end{abstract}

\section{Introduction}\label{sec-introduction}

\begin{figure*}
    \centering
    \includegraphics[width=\linewidth]{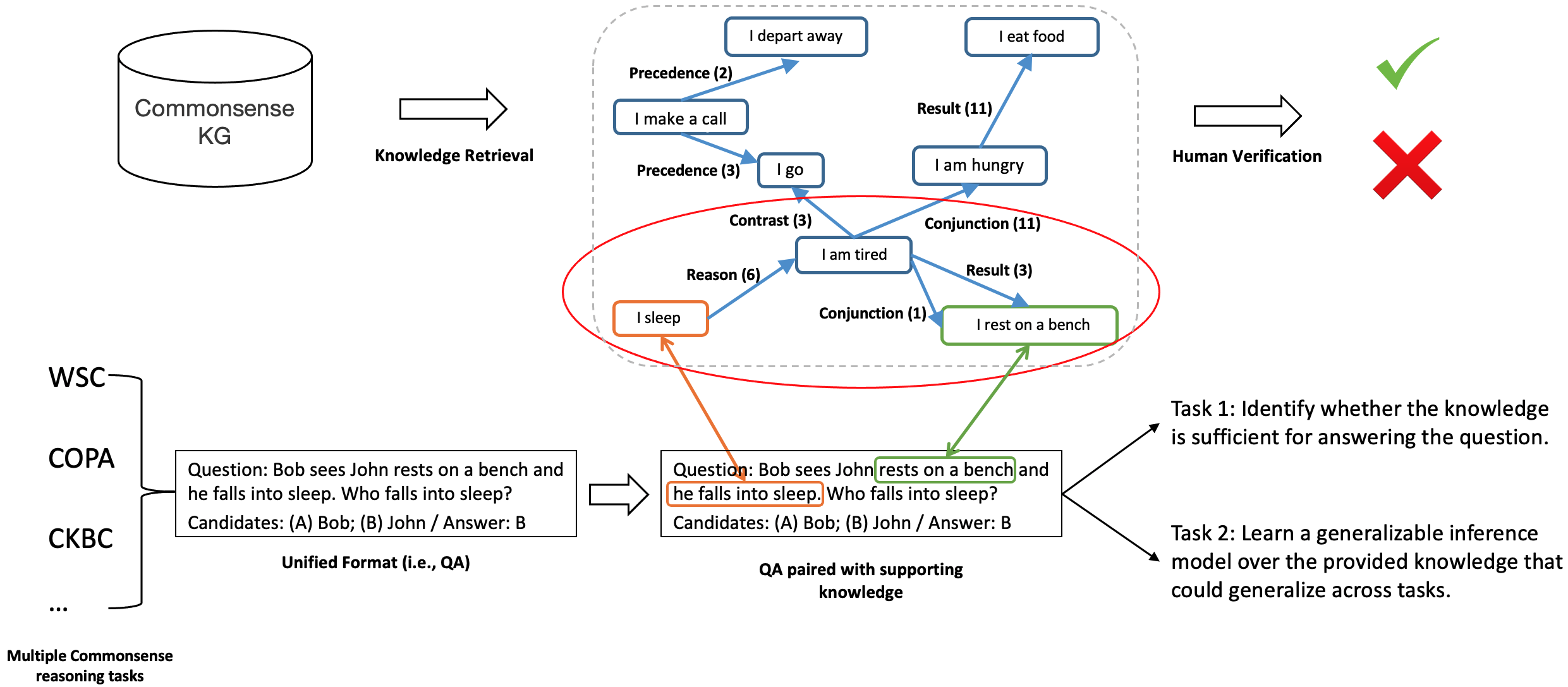}
    \caption{\name~ demonstration. All tasks are converted into a unified format such that we could easily evaluate the generlization capability of all models. We also equip all questions with auto-extracted knowledge graphs from existing KGs and ask humans to annotate whether the knowledge is gold or not. In this example, we expect models to first identify the quality of the knowledge and then conduct inference over the knowledge to solve the question. }
    \label{fig:intro_demo}
    \vspace{-0.2in}
\end{figure*}


Understanding human language requires both the language knowledge (e.g., grammar and semantics) and world knowledge, which can be further divided into factual and commonsense knowledge \cite{Katz1963-KATTSO-3}.
Recently, the community has made great progress on helping machines acquire and apply language and factual knowledge. 
However, how to help machines acquire and infer over commonsense is still unclear. 
To answer this question, many commonsense reasoning datasets~\cite{DBLP:conf/aaaiss/RoemmeleBG11,DBLP:conf/aaai/SakaguchiBBC20,DBLP:conf/naacl/TalmorHLB19,DBLP:conf/cvpr/ZellersBFC19,DBLP:conf/emnlp/LinLKR20} have been proposed. Even though they target different knowledge types, modalities, and come in different formats, they often follow a standard supervised learning setting, which aims at helping machines to solve a specific task with the training data.
However, two limitations of this learning paradigm have restricted the development of commonsense reasoning systems.

First, there is no clear separation between knowledge and inference. As discussed in~\cite{DBLP:journals/corr/abs-2104-08161}, a common phenomenon is that larger training data will lead to better performance, mainly because richer knowledge is covered. However, due to the large scale of commonsense knowledge, it is infeasible to annotate a large enough training set for each task, and the responsibility of the training data should be teaching models how to do inference rather than acquire the commonsense knowledge. 
Several recent works have explored using structured knowledge for commonsense reasoning tasks~\cite{DBLP:conf/emnlp/LinCCR19,DBLP:conf/aaai/LvGXTDGSJCH20,DBLP:conf/emnlp/PaulF20}. However, as these works did not clearly analyze the coverage of the structured knowledge (i.e., knowledge graphs (KGs)), it is still unclear what the performance means, better knowledge coverage or better inference capability. 
To dig into what is behind this learning process, we propose to equip each question with auto-extracted knowledge and ask humans to annotate whether the knowledge is gold (i.e., sufficient to answer the question).
By doing so, we could evaluate whether models can know if the provided knowledge is gold or not and how well they can conduct inference over the provided knowledge to solve the task.

Second, the supervised learning may force the model to learn the distribution of the training data rather than a universal inference model. As a result, the model may perform well on the test set that follows the same distribution but fail on other tasks~\cite{DBLP:journals/corr/abs-2011-09159}.
Previously, as different tasks have different formats, it is hard to evaluate the generalization ability of commonsense reasoning models.
Motivated by the existing trend of using a unified format (i.e., question answering) for different tasks~\cite{DBLP:conf/emnlp/KhashabiMKSTCH20}, we propose to convert various commonsense reasoning tasks into a unified QA format such that we can easily and fairly evaluate the generalization ability of learned commonsense reasoning models.

Combining these two lines of effort, we propose a new commonsense inference evaluation benchmark Knowledge-in-the-loop Commonsense Inference with QA (\name).
An example is shown in Figure~\ref{fig:intro_demo}. We first convert several popular commonsense reasoning tasks into a unified QA format and equip them with the relevant knowledge from existing commonsense knowledge graphs.
We leverage human annotation to label whether the provided knowledge is gold to answer the question.
With \name, we are interested in answering two questions: (1) Whether current models can distinguish the knowledge is gold or not; 
(3) Can current commonsense inference models generalize across different commonsense reasoning tasks.

Experiments with several recent knowledge-based commonsense reasoning models show that even though current deep models could learn to conduct simple inference after training with a few examples when gold knowledge is provided, they still cannot learn to distinguish gold knowledge very well.
Moreover, even though current models demonstrate an encouraging generalization ability across the three tasks we consider, they still cannot learn complex inference (e.g., abductive reasoning) very well.
We hope that our benchmark\footnote{Available at https://github.com/CogComp/CIKQA.} can motivate more advanced commonsense inference methods in the future.



\section{Dataset Construction}\label{sec:definition}

In \name, to encourage a generalizable commonsense inference model, we follow previous work~\cite{DBLP:conf/emnlp/KhashabiMKSTCH20,DBLP:journals/corr/abs-2010-04829,DBLP:conf/acl/WuWYWL20,DBLP:conf/emnlp/DuC20} to unify all selected tasks as a binary question answering problem, and equip each question with a supporting knowledge graph $G$ retrieved from existing commonsense KGs.
We leverage crowd-sourcing workers to annotate whether the knowledge is gold (i.e., accurate and enough) for answering the question.
Details about task selection, format unification, support knowledge extraction, and annotation are as follows.

\begin{table*}[t]
    \small
    \centering
    \begin{tabular}{l||p{4.0cm}|p{4.5cm}|p{3.5cm}}
    \toprule
        Task Name & Original Assertion & Transformed Question &  Answer \\
    \midrule
       HardPCR  & The fish ate the worm. It was hungry. & The fish ate the worm. It was hungry. What was hungry? & {(A) \Blue{Fish}; (B) \Red{Worm}} \\
       \hline
       CommonsenesQA  & What is a place that someone can go buy a teddy bear? & What is a place that someone can go buy a teddy bear? & (A) \Blue{Toy store}; (B) \Red{Shelf}\\
       \hline
       COPA  & I drank from the water fountain. & I drank from the water fountain. What was the cause of this? & (A) \Blue{I was thirsty.}; (B) \Red{I felt nauseous.} \\
       \hline
       ATOMIC & PersonX buys the bike. & Before PersonX buys the bike, what did PersonX want? & (A) \Red{To be social.}; (B) \Blue{To have transportation.}\\
    \bottomrule
    \end{tabular}
    \caption{Demonstration of the original assertion, transformed questions, and answers. Correct and wrong answers are indicated with blue and red, respectively.}
    \vspace{-0.1in}
    \label{tab:Commonsense_Task_Demonstration}
\end{table*}

\subsection{Task Selection}\label{sec:task_selection}
In \name, we select the following four popular commonsense reasoning tasks:
\begin{enumerate}[leftmargin=*]
        \item HardPCR~\cite{DBLP:journals/corr/abs-2009-12721}: The hard pronoun coreference resolution (HardPCR) task is one of the most famous commonsense reasoning tasks. For each question, a target pronoun and two candidate mentions are provided, and the task is to select the correct mention that the pronoun refers to. Careful expert annotations are conducted to get rid of the influence of all simple linguistic rules and the models are required to solve the problem with commonsense reasoning. In \name, we include instances from WSC~\cite{levesque2012winograd}, DPR~\cite{DBLP:conf/emnlp/RahmanN12}, and WinoGrande~\cite{DBLP:conf/aaai/SakaguchiBBC20}. 
        To create a question regarding the target pronoun, we first find the sentence that contains the target pronoun and then determine whether the participating pronoun refers to a person or an object.

    \item CommonsenseQA~\cite{DBLP:conf/naacl/TalmorHLB19}: CommonsenseQA is a commonsense question answering dataset. For each question-answer pair, four relevant but wrong concepts are used as the other candidates, and the models are required to select the correct one out of five candidates. In \name, we randomly sample a negative answer to make it a binary choice task, which is consistent with other datasets. 
    \item COPA~\cite{DBLP:conf/aaaiss/RoemmeleBG11}: COPA focuses on evaluating the understanding of events causality. For a target event, two candidate followup events are provided, and models are asked to predict the one caused by or the reason for the target event.
    
    \item ATOMIC~\cite{sap2019atomic}: The last one is the commonsense knowledge base completion. Given a head concept (e.g., ``eat food'') and a relation (e.g., ``cause''), we want to predict the tail concept. In \name, we focus on predicting edges of ATOMIC.

\end{enumerate}

In COPA and ATOMIC, where the task is to predict the relations between two events or states (e.g., ``PersonX eats''-\textit{Causes}-``PersonX is full''), for each triplet, we randomly sample another event or state as the negative tail and ask the model to select the correct one. 
To make the task challenging and avoid sampling irrelevant events or states, we require the sampled negative event or state to be connected with the head event or state with a different triplet (e.g., ``PersonX is hungry'' from the triplet ``PersonX eats''-\textit{CausedBy}-``PersonX is hungry''). 
For each type of relation, we write a pattern to generate the question. For example, for the ``Causes'' relation, we will ask ``What can be caused by `PersonX eats'?''. 
Examples of instances in the original datasets and their transformed questions and candidate answers are presented in Table~\ref{tab:Commonsense_Task_Demonstration}.

\subsection{Supporting Knowledge Extraction}\label{sec:knowledge_extraction}


As discussed in Section~\ref{sec-introduction}, a limitation of existing commonsense reasoning benchmarks is that there is no clear boundary between knowledge and inference. As such, it is unclear what is learned from the training data, the knowledge, or how to perform inference.
To address this issue and encourage models to learn inference rather than knowledge from the training data, we propose to equip each question with supporting knowledge. 
The question is selected as part of the dataset only if we find supporting knowledge to answer the question.
Note that this procedure serves as an improved evaluation setup than pure supervised learning, and not as a solution to commonsense reasoning.
This section introduces the selected commonsense knowledge graphs and then introduces how we extract the corresponding commonsense knowledge for each question.

\subsubsection{Commonsense KG Selection}

Many commonsense knowledge graphs were developed to enhance machines' commonsense reasoning abilities, including ConceptNet~\cite{liu2004conceptnet}, ATOMIC~\cite{sap2019atomic}, GLUCOSE~\cite{mostafazadeh-etal-2020-glucose}, and ASER~\cite{zhang2019aser}.
Among these four, ConceptNet, ATOMIC, and GLUCOSE were constructed via crowd-sourcing while ASER was constructed automatically with information extraction techniques.
Besides ATOMIC, which is used as one of the tasks, we use the other KBs as supporting knowledge resources.

\subsubsection{Supporting Graph Extraction}

Here we introduce how to extract the supporting knowledge from external commonsense knowledge bases.
For each question, we need to obtain a sub-graph from supporting knowledge graphs such that it contains the relevant commonsense knowledge about the question. The sub-graph extraction process includes the following three steps: (1) Pre-processing: Convert each question into several key sentences; (2) Matching: Match the sentences into nodes in the KG; (3) Extraction: Retrieve the relevant sub-graphs from the KG. 

\noindent \textbf{Data Pre-processing}: For each question and the associated candidate answers, we first replace the question words (e.g., ``What'') with the two candidate answers such that it becomes two declarative sentences.
For instance, if the question is ``The fish ate the worm. It was hungry. Who is hungry?'' and the candidates are ``Fish'' and ``Worm,'' we will convert the question into the declarative sentence: ``The fish is hungry'' and ``The worm is hungry.'' 
As a result, we will get three sentences for this question: ``The fish ate the worm,'' ``The fish is hungry,'' and ``The worm is hungry.''

\begin{table*}[t]
    \small
    \centering
    \begin{tabular}{l||c|c|c||c|c|c}
    \toprule
        \multirow{2}{*}{Task Name} & \multicolumn{3}{c||}{\# Instance by Knowledge Resource} & \multirow{2}{*}{\# Total Instance}& \multirow{2}{*}{Avg Sub-graph Size} & \multirow{2}{*}{\# Gold Instance} \\
        & ASER & ConceptNet & GLUCOSE  &  & & \\
    \midrule
   HardPCR & 2,030 & 202 & 2,143 & 4,375 & 2.85 & 670 \\
   CommonsenseQA  & 530 & 31 & 37 & 598 & 3.19 & 59\\
   COPA & 103 & 41 & 149 & 293 & 3.03 & 78\\
   ATOMIC  & 5,655 & 212 & 3,466 & 9,333 & 2.67 & 2,200\\
       \midrule
       Total & 8,318 & 486 & 5,795 & 14,599& 2.75 & 3,007\\
    \bottomrule
    \end{tabular}
    \caption{\name ~statistics. ``Avg Sub-graph Size'' is the average graph size, which is measured by the number of edges. ``\# Gold Instance'' means the number of instances supported by different knowledge resources and annotated gold (i.e., Accurate and Enough) knowledge. }
    \label{tab:dataset_statistics}
    \vspace{-0.2in}
\end{table*}

\noindent \textbf{KG Matching}: After getting the declarative sentences that contain the question and key answers, to extract the relevant knowledge, we map them to nodes in knowledge graphs. Considering that each sentence may have multiple words and it is often hard to find an exact match, we adopt an embedding-based fuzzy matching technique. 
For each sentence and node in the KG, we treat them as a sentence and get the corresponding representations with SimCSE~\cite{DBLP:conf/emnlp/GaoYC21}. For each input sentence, SimCSE encodes the sentence into a vector. A close distance between two vectors indicates that the two sentences are similar to each other.
We use cosine similarity on the obtained representations to measure the similarity between two sentences.\footnote{We also tried other techniques such as string match, ROUGE~\cite{lin2004rouge}, and BLEURT~\cite{DBLP:conf/acl/SellamDP20}, but found them to be either inaccurate or too slow for our scale.}
Since there are 287 thousand nodes in GLUCOSE and 194 million nodes in ASER, it is computationally infeasible to compute the cosine similarity between sentences pair by pair.
Thus we use an approximation. 
For each extracted sentence, we first apply Faiss~\cite{DBLP:journals/corr/JohnsonDJ17}, a large-scale similarity-based matching algorithm that first clusters all KG nodes in the vector space to increase the matching efficiency when finding the top $N$ nodes in the KG. 
We encode all the nodes of the graph and index them using Faiss~\cite{DBLP:journals/corr/JohnsonDJ17}. Then, we can perform fast and quick retrieval of the most-similar nodes with each query sentence.
After that, we sort the $N$ nodes based on the cosine similarity to find the top $K$ similar nodes.
We set $N$ and $K$ to be 60 and 1, respectively.
On average, it takes 25 seconds to retrieve the relevant nodes for each question.

\noindent \textbf{Graph Extraction}: Next, we extract the sub-graph that contains all the relevant nodes. We denote the extracted $m$ nodes as $n_1, n_2, ..., n_m$, and for each of them, we find $K$ similar nodes from KG. The resulting matched node sets are denoted as $\NM_1, \NM_2, ..., \NM_m$. For any pair of nodes $n \in \NM_i$ and $n^\prime \in \NM_j$ ($i \neq j$), if there exist a path in the KG between $n$ and $n^\prime$, we will keep that path. After adding all paths together, we will get the final sub-graph.
On average, it takes less than two seconds to construct a graph for each question.

\noindent \textbf{Knowledge Quality Annotation}: 
Since our extraction method is an automatic one, some of the subgraphs may be irrelevant or insufficient for answering the questions. We use crowdsourcing to annotate whether the extracted knowledge is gold (i.e., accurate and enough). For each question, we invite five annotators to provide the annotation. The average Inter-annotator agreement (Cohen’s kappa statistic) is 0.83, which indicates the high-quality of our annotation. In the end, we apply a strict standard (at least four of five annotators need to vote for gold) to select the gold knowledge.
More annotation details could be found in Appendix Section~\ref{sec:annotation}.

\subsection{\name~ Statistics}

We report the dataset statistics in Table~\ref{tab:dataset_statistics}. 
In total, we collect 14,599 instances, and among which Hard PCR and ATOMIC provide the most questions because their original datasets are much larger than others.
According to the annotation, 16.69\% of the supporting knowledge graphs are gold knowledge. 
Based on our analysis, annotators hold a very strict standard for selecting the gold knowledge.
For each task, we randomly split the dataset into training, development, and testing set with a standard 8:1:1 splitting.
As a result, we get 11,678 training, 1,459 development, and 1,462 testing instances.

\section{Experiment Setup}\label{sec:experiment}

We present the performance of following commonsense inference models on \name:

\noindent \textbf{(1) Vanilla LM}: We use the language model (LM) based multiple-choice (MC) model as the basic baseline. For each candidate answer, we concatenate it with the question and feed it to the model. After getting the sentence representation, a linear layer is used to obtain a score and trained with a cross-entropy loss. 

    \noindent \textbf{(2) KagNet}: As one of the pioneering works that utilized structured knowledge for solving commonsense reasoning tasks, KagNet~\cite{DBLP:conf/emnlp/LinCCR19} first uses a graph convolution network to encode the knowledge graph and then apply an LSTM based hierarchical attention mechanism to encode the knowledge paths that start with the nodes corresponding to the question and end with nodes corresponding to the answer. At the same time, KagNet encodes the question and answers with pre-trained LMs. In the end, it concatenates all representations for the final prediction. 
    
    \noindent \textbf{(3) Graph Based Reasoning (GBR)}: Instead of only encoding paths starting with the question nodes and ending with answer nodes, in GBR~\cite{DBLP:conf/aaai/LvGXTDGSJCH20}, they proposes to run a depth-first algorithm over the knowledge graph to generate a sequence of paths as the supporting knowledge paths.
    
    \noindent \textbf{(4) Multi-Head Knowledge Attention (MHKA)}: To further utilize the knowledge, MHKA~\cite{DBLP:conf/emnlp/PaulF20} uses a transformer network to model the paths from the question nodes and answer nodes, then concatenates the knowledge and context representation for the final prediction. 
    
    \noindent \textbf{(5) Graph-to-Text (G2T)}: In the end, we also evaluate a simple yet effective approach of combining structured knowledge and language models: Graph-to-Text~\cite{DBLP:conf/aaai/BianH0021}, which first verbalizes knowledge into a sentence and then concatenates the knowledge sentence and target question together. On top of that, a transformer-based model is used to encode the input the sentence and make the final prediction.

\paragraph{Implementation Details}
We implement all experiments with Huggingface~\cite{DBLP:journals/corr/abs-1910-03771}.
We select BERT-base
~\cite{DBLP:conf/naacl/DevlinCLT19} as the base language model for all models. The batch size is set to be 16. All models are trained for 10,000 steps\footnote{All models converge at 10,000 steps.}, and the best-performing checkpoints on the dev set are evaluated.
For our model, we set both the number of random walk paths and walk length to be five.
Considering that the auto-extracted knowledge could contain noise or miss certain knowledge, we add a ``gold knowledge'' setting, where only examples with the gold knowledge are used for training and testing, for all models as the upper bound of their model.
All other hyper-parameters are the same as the base language model. 
All models are trained with GTX 2080 and the average running time is 12 hours.

\section{Result Analysis}

\begin{figure}
    \centering
    \includegraphics[width=0.8\linewidth]{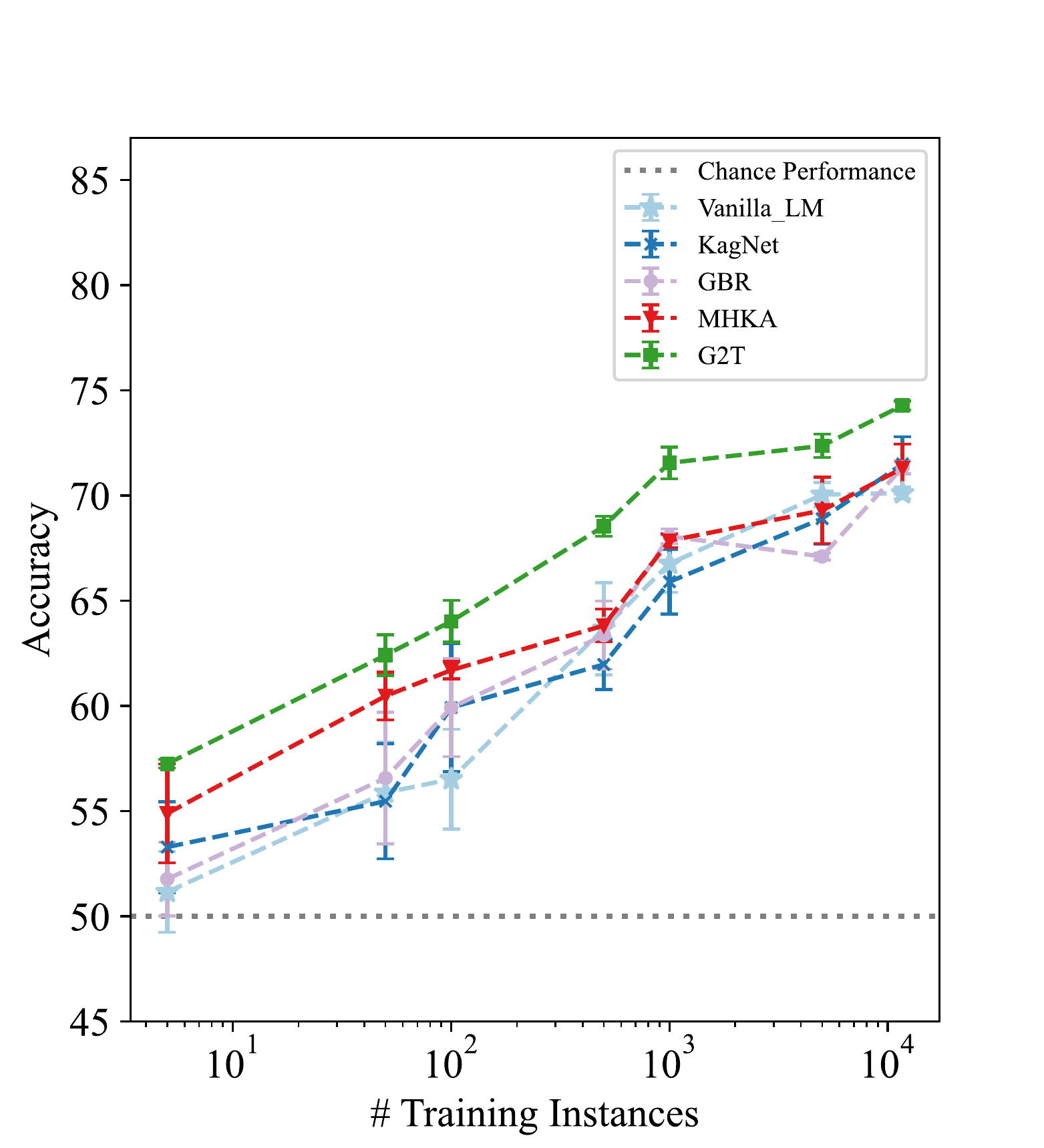}
    \caption{Learning curves of all evaluated models on all instances of \name. }
    \label{fig:all_instances}
\end{figure}

\begin{figure}[t]
    \centering
    \includegraphics[width=0.8\linewidth]{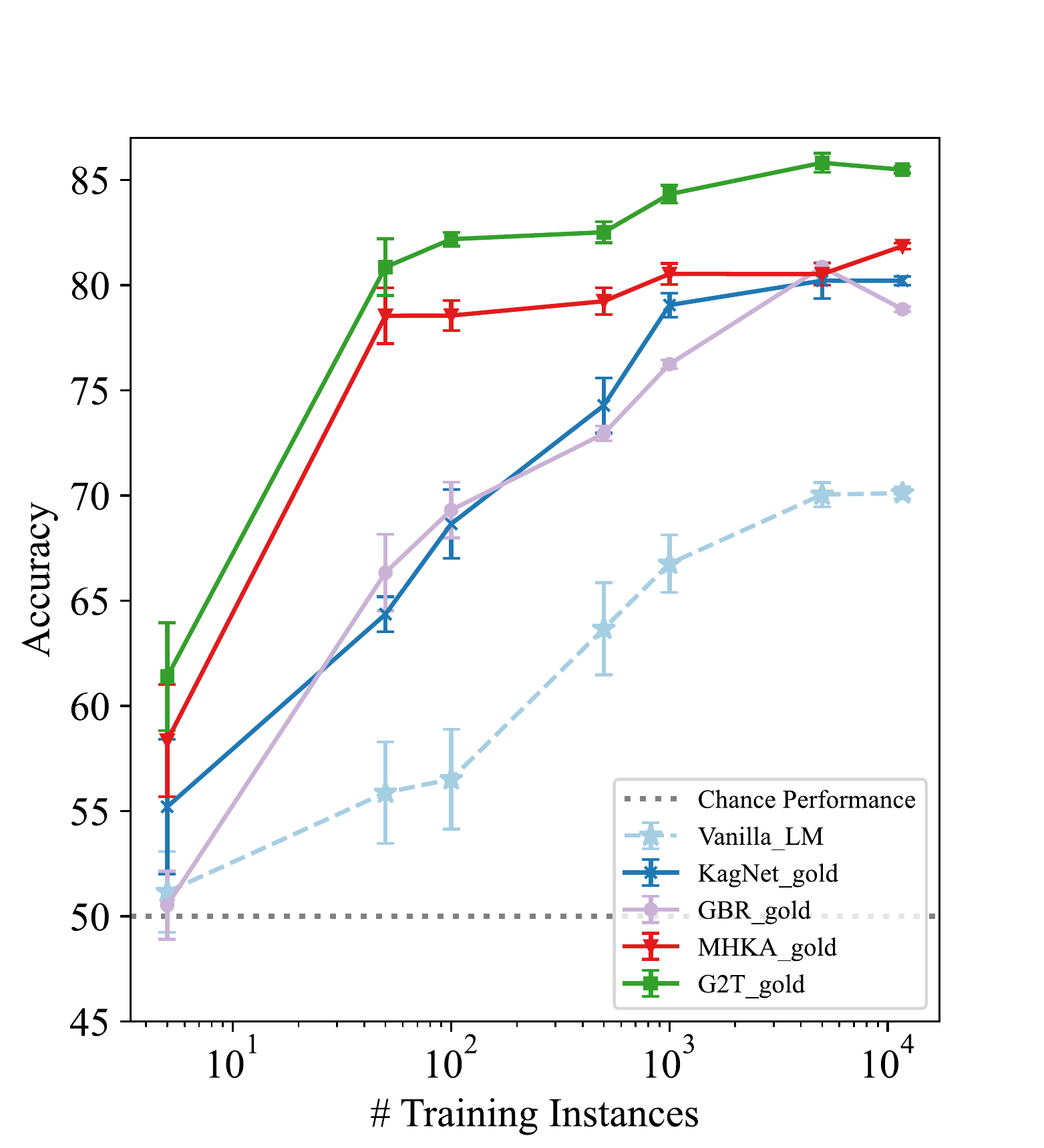}
    \caption{Learning curves of all evaluated models on the gold subset of \name, where only instances with gold knowledge are used for training and testing. }
    \label{fig:gold_instance}
    \vspace{-0.1in}
\end{figure}

We first conduct analysis experiments to evaluate to what extent the provided knowledge could help existing models. 
For each model, we train it with different numbers of training instances and report the average performance and standard deviation\footnote{Due to the space limitation, we put the detailed experimental results in Appendix Section~\ref{sec:detailed_experimental_results}.} of five trails.
Experiment results with all instances and the gold subset of \name, where only instances with gold knowledge are used for training and testing, are presented in Figure~\ref{fig:all_instances} and~\ref{fig:gold_instance}, respectively.
From the results, we can make the following observations. First, when explicitly including the knowledge, all inference models outperform the baseline model that has no support of the knowledge, especially G2T. When the auto-extracted knowledge and gold knowledge are provided, G2T outperforms the baseline Vanilla LM model by 4.17 and 15.34 accuracy, respectively.
It supports our assumption that it is hard to learn all knowledge from the limited training data and external structured knowledge could help.
At the same time, we also notice that there is a significant gap between auto-extracted knowledge and gold knowledge. For example, models could learn to answer the questions with only a small number of examples if gold knowledge is available. 
This indicates that the knowledge quality can significantly impact models' performance, which further shows the importance of distinguishing whether the knowledge is gold or not automatically.
Last but not least, we can see that G2T outperforms other inference models among most settings, which shows that with the help of current large-scale LMs, jointly encoding question and knowledge is more efficient and a more effective strategy than acquiring them separately. Due to the simplicity and efficiency of G2T, we will conduct the rest analysis experiments with G2T.

\subsection{Distinguishing the Gold Knowledge}

\begin{table*}[t]
	\centering
    \small
    \vspace{-0.05in}
    \subtable[Full Dataset (Vanilla LM (without knowledge)$\rightarrow$ G2T (with knowledge))]{
  \begin{tabular}{l||c|c|c|c}
    \toprule
     \multirow{2}{*}{Training Task}                  & \multicolumn{4}{c}{Testing Task}\\
 \cline{2-5}
    &Hard PCR  & CommonsenseQA & COPA  & ATOMIC \\
         \midrule
    Hard PCR   &  -  &  37.50 $\rightarrow$ 52.30  & 75.00 $\rightarrow$ 53.24  &  44.13 $\rightarrow$ 53.32  \\
    CommonsenseQA   &  50.00 $\rightarrow$ 50.14  &  -  &  62.50 $\rightarrow$ 56.67  &  56.34 $\rightarrow$ 70.56  \\
    COPA    &  45.95 $\rightarrow$ 51.26  &  62.50 $\rightarrow$ 58.33 &  -  &  49.77 $\rightarrow$ 62.96  \\
    ATOMIC   &  39.19 $\rightarrow$ 50.76  &  50.00 $\rightarrow$ 76.67  &  62.50 $\rightarrow$ 73.33  &  -  \\
    \bottomrule
    \end{tabular}
	}

    \subtable[Gold Subset (Vanilla LM (without knowledge)$\rightarrow$ G2T (with knowledge)) ]{
    \begin{tabular}{l||c|c|c|c}
    \toprule
     \multirow{2}{*}{Training Task}                  & \multicolumn{4}{c}{Testing Task}\\
 \cline{2-5}
    &Hard PCR  & CommonsenseQA & COPA  & ATOMIC \\
         \midrule
    Hard PCR  &  -  &  46.67 $\rightarrow$ 51.67  &  63.33 $\rightarrow$ 56.67  &  51.85 $\rightarrow$ 55.78  \\
    CommonsenseQA   &  49.32 $\rightarrow$ 50.32  & - & \hlc[orange]{ 50.00 $\rightarrow$ 75.00 } & \hlc[green]{ 60.39 $\rightarrow$ 91.08 }\\
    COPA     &  52.51 $\rightarrow$ 54.79  & \hlc[orange]{ 56.67 $\rightarrow$ 87.50 } &  -  & \hlc[green]{ 53.01 $\rightarrow$ 76.06 }\\
    ATOMIC   &  50.46 $\rightarrow$ 51.35  & \hlc[green]{ 68.33 $\rightarrow$ 93.75 } & \hlc[green]{ 56.67 $\rightarrow$ 87.50 } &  - \\
    \bottomrule
    \end{tabular}
	}
	\vspace{-0.1in}
	\caption{Generalization ability demonstration. 
	We report the performance on both the full dataset and gold dataset (i.e., only questions with gold knowledge are selected for training and testing)  to show the generalization ability. Strong and moderate generalization settings are indicated with the  \hlc[green]{green} and \hlc[orange]{orange} background, respectively.} 
\label{tab:Generalization_ability}
\end{table*}

\begin{figure}[t]
    \centering
    \includegraphics[width=0.8\linewidth]{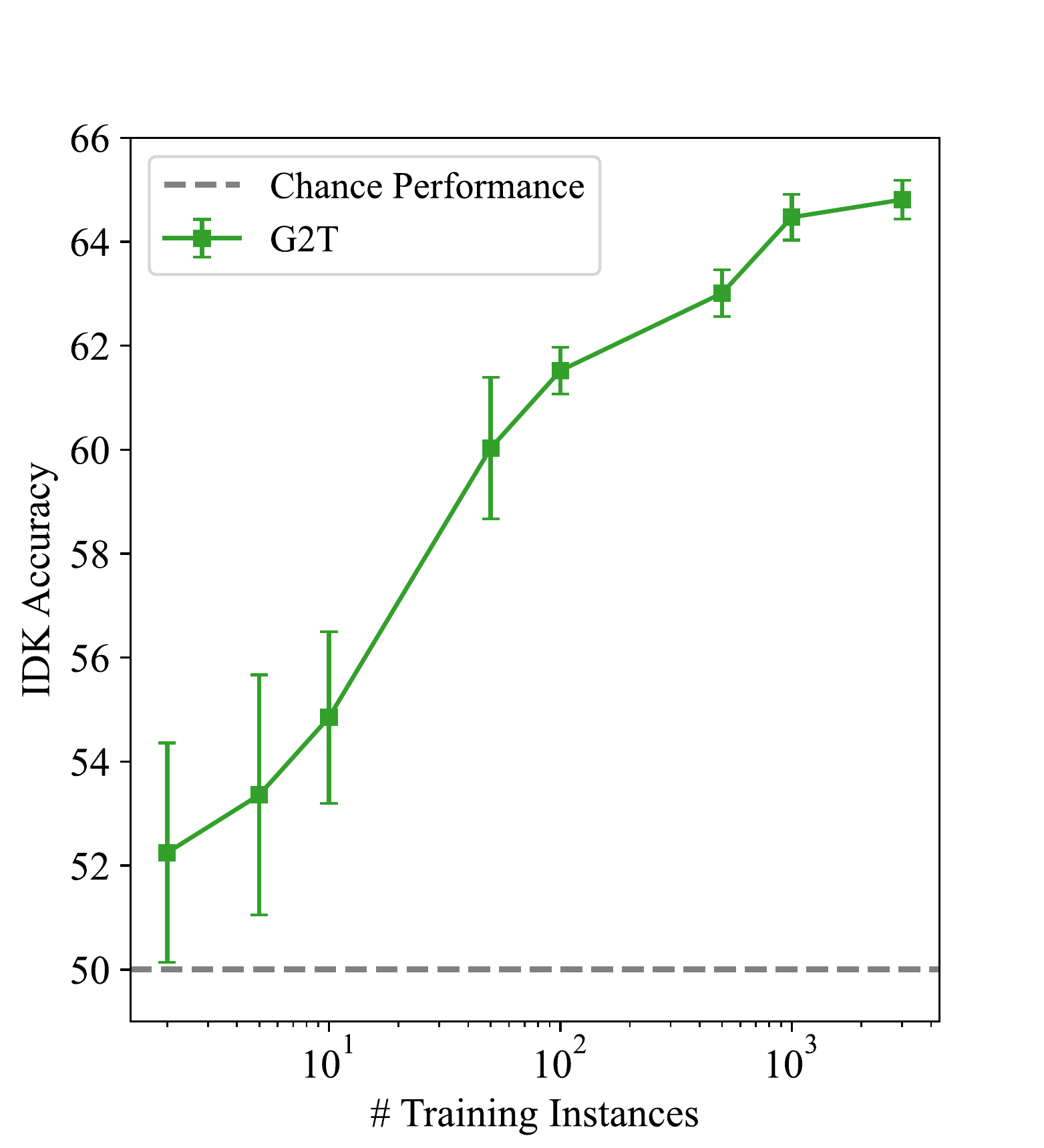}
    \caption{The learning curve of G2T on the gold knowledge identification task.}
    \vspace{-0.2in}
    \label{fig:IDK_results}
\end{figure}

Humans have the capability of saying ``I do not know'' when they find out that they cannot answer a question with their knowledge.
To investigate whether current deep models have a similar capability, we use G2T as an example to test whether these deep models can distinguish the gold knowledge.
For each (question, answer, and knowledge) triplet, we train and test G2T with annotated knowledge quality labels.
To address the imbalanced distribution problem, we randomly 
select the same number of ``Not Gold'' examples as the ``Gold'' ones to make the dataset balanced.
From the results in Figure~\ref{fig:IDK_results}, we can see that the performance of G2T can be improved  slightly with the increase of training data.
However, after seeing thousands of examples, it still can only achieve 0.65 accuracy on a binary classification problem.
It shows that knowing when to say ``I do not know'' is still a challenging task for current deep models, which is consistent with the observations in previous literature that deep models cannot understand the reasons and knowledge they used to answer questions~\cite{DBLP:conf/acl/ZhangZS20,DBLP:journals/corr/abs-2110-08207}.
We hope that \name~could motivate more future work on this important research problem.

\subsection{Generalization Ability}

An important assumption and motivation behind the unified problem design of \name~is that even though the commonsense could be enormous, the inference rules over commonsense knowledge can be limited.
As a result, even though we could not learn all the commonsense from limited training data, we can learn how to conduct inference with several tasks and then generalize to others.
In this section, we conduct experiments with both the ``Without Knowledge'' and ``With Knowledge'' models to show that with our unified formulation, we can gain such generalization ability across different tasks.
We conduct experiments on two settings: (1) Full Set: We train and test the model with the whole dataset; (2) Gold Subset: We only train and test the model on questions, where the supporting graph is annotated as gold.
We train the model with questions from a specific task and test it on all tasks. The results are in Table~\ref{tab:Generalization_ability}.

\begin{figure*}
    \centering
    \includegraphics[width=0.95\linewidth]{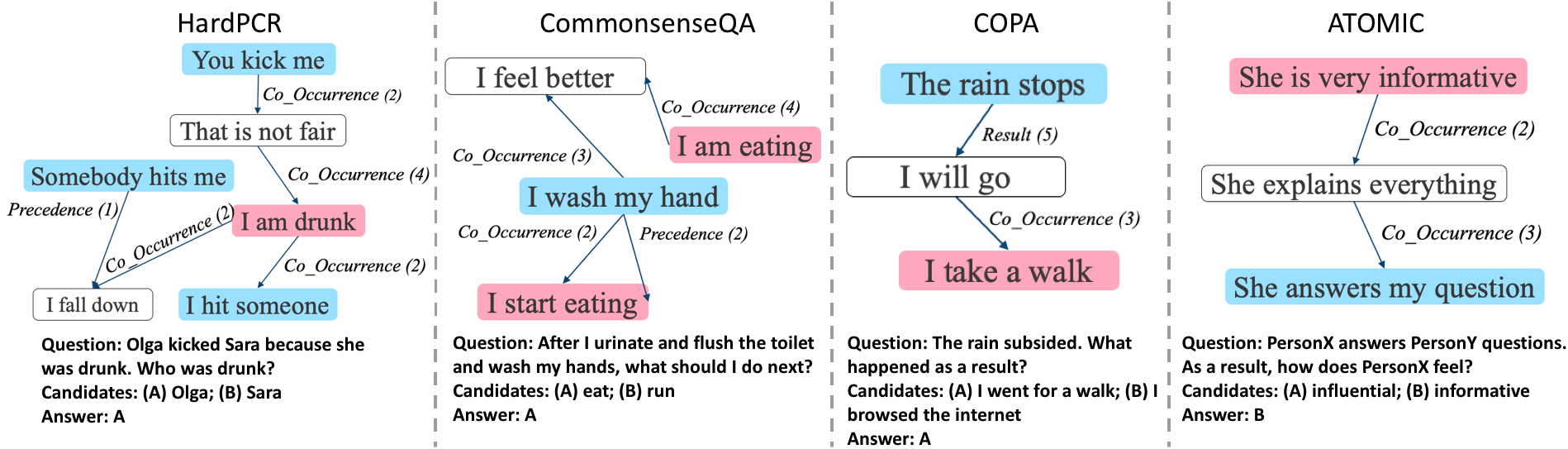}
    \vspace{-0.01in}
    \caption{\name~ Case Study. Mapped nodes for the question/answers are in blue/pink. Other nodes are white. Edge weights are in brackets. We only show the relevant parts of the graphs for clear representation. }
    \vspace{-0.1in}
    \label{fig:case_study}
\end{figure*}

From the results, we can see that the knowledge can help models to generalize well among CommonsenseQA, COPA, and ATOMIC. The only exception is HardPCR.
This is mainly because the inference needed for solving HardPCR is more complex than the other tasks, where we do not only need to find the relevant knowledge but also need to replace the target pronouns with the entity in the provided knowledge.
As shown in Figure~\ref{fig:case_study}, two paths can be found relevant to question: (1) ``I am drunk''$\rightarrow$\textit{Co\_Occurrence}$\rightarrow$``I hit someone''; (2) ``I am drunk''$\rightarrow$\textit{Co\_Occurrence}$\rightarrow$``That is not fair''$\rightarrow$\textit{Co\_Occurrence}$\rightarrow$``You kick me''. For the correct inference, we need to know when there is a conflict, we should trust the one-hop inference more because the additional node in the two-hop path may introduce extra noise.
As a comparison, for other tasks, the main inference we need is to find the relevant paths, which is relatively easy.
How to train a model that can learn to conduct such complex reasoning is a problem worth exploring in the future.

In general, the observed generalization ability is encouraging because if we can learn a good model on \name, based on the assumption that there are limited types of inference, potentially we can solve any commonsense reasoning task as long as the needed inference types are covered by \name. At the same time, we also notice that models typically generate better when gold knowledge is provided, which further proves the importance of the gold knowledge identification task.








\section{Related Work}\label{sec:related_works}



To help machines understand commonsense, the community has devoted great efforts in constructing commonsense knowledge bases with either crowdsourcing (e.g., ConceptNet~\cite{liu2004conceptnet} and ATOMIC~\cite{sap2019atomic}) or information extraction techniques (e.g., ASER~\cite{zhang2019aser}).
Typically, crowd-sourced knowledge bases are of higher quality, and the auto-constructed ones have larger coverage.
Besides acquiring commonsense knowledge, the community also developed many commonsense reasoning datasets to train and test models' commonsense reasoning abilities. Even though these datasets may have different \textit{formats} (e.g., slot fitting in Winogrande~\cite{DBLP:conf/aaai/SakaguchiBBC20} and question answering in CommonsenseQA~\cite{DBLP:conf/naacl/TalmorHLB19}), \textit{knowledge types} (e.g., causal commonsense in COPA~\cite{DBLP:conf/aaaiss/RoemmeleBG11} and numerical commonsense in NumerSense~\cite{DBLP:conf/emnlp/LinLKR20}), or \textit{modalities} (e.g, visual commonsense in VCR~\cite{DBLP:conf/cvpr/ZellersBFC19} and textual commonsense in many others), they follow a standard supervised learning setting, and aim at helping machines to solve a specific commonsense task in an end-to-end manner.
Given this setting, it is often difficult to tell what has been learned during the training process. Was it used to acquire commonsense knowledge, learn to conduct commonsense inference, or both?
Such ambiguity limits our progress in solving these commonsense reasoning tasks.
In this work, we connect the efforts on commonsense acquisition and inference by creating a commonsense inference benchmark \name~, where models can focus on learning to identify the gold knowledge and perform inference over the supporting commonsense knowledge.

Answering questions in natural language based on a knowledge base (KB) is a mature research topic in the NLP community, which is also known as the KBQA problem~\cite{clark1999knowledge,DBLP:conf/acl/YihCHG15,DBLP:conf/acl/YihRMCS16,DBLP:conf/esws/UsbeckNHKRN17,DBLP:journals/pvldb/CuiXWSHW17}.
Previous work mainly focuses on factual knowledge, which is stored in the format of triplets, and the main challenge is to parse the question and then precisely and effectively identify the correct path over a large-scale KB to do the inference.
Compared with inference over factual knowledge, inference over commonsense knowledge brings the following unique challenges: 
(1) Commonsense is a kind of preference rather than fixed knowledge. As a result, the ideal commonsense reasoning process could involve the comparison of multiple candidates . For example, both ``drink coffee'' and ``drink bear'' could happen in the morning, but a normal person will prefer ``drink coffee;''
(2) Beyond named entities, commonsense knowledge also covers daily entities and events, and thus it is difficult to find an exact node from the commonsense KB that matches the question and we may need to conduct inference based on the partial match (i.e., the extracted nodes are relevant but not identical).

\section{Conclusion}\label{sec:conclusion}

In this paper, we present \name, a unified commonsense inference benchmark. 
Specifically, we first convert several popular commonsense tasks into a unified QA format and then equip each question with a supporting commonsense knowledge graph.
We also leverage humans to annotate the quality of auto-extracted knowledge.
Experiments show that even though models can better learn how to do commonsense inference with a few examples and significantly outperform the baseline method that does not use structured knowledge in the data-scarce setting, how to identify the gold knowledge is still an unsolved problem.
More interestingly, with our unified formulation, models demonstrate the encouraging generalization ability across tasks.
As both the format unification and supporting graph extraction are automatic, we can easily extend to other commonsense reasoning tasks in the future. 
All used code and data are submitted in the submission system.

\section*{Acknowledgements}

The authors of this paper were supported by the Office of the Director of National Intelligence (ODNI), Intelligence Advanced Research Projects Activity (IARPA), via IARPA Contract No. 2019-19051600006 under the BETTER Program, and by contract FA8750-19-2-1004 with the US Defense Advanced Research Projects Agency (DARPA).
The views expressed are those of the authors and do not reflect the official policy or position of the Department of Defense or the U.S. Government.
This paper was also supported by the NSFC Fund (U20B2053) from the NSFC of China, the RIF (R6020-19 and R6021-20) and the GRF (16211520) from RGC of Hong Kong, the MHKJFS (MHP/001/19) from ITC of Hong Kong with special thanks to HKMAAC and CUSBLT, and  the Jiangsu Province Science and Technology Collaboration Fund (BZ2021065).
Yanai Elazar is grateful to be supported by the PBC fellowship for outstanding PhD candidates in Data Science and the Google PhD fellowship.


%





\bibliography{main}

\begin{thebibliography}{36}
\expandafter\ifx\csname natexlab\endcsname\relax\def\natexlab#1{#1}\fi

\bibitem[{Bian et~al.(2021)Bian, Han, Chen, and Sun}]{DBLP:conf/aaai/BianH0021}
Ning Bian, Xianpei Han, Bo~Chen, and Le~Sun. 2021.
\newblock \href {https://ojs.aaai.org/index.php/AAAI/article/view/17490}
  {Benchmarking knowledge-enhanced commonsense question answering via
  knowledge-to-text transformation}.
\newblock In \emph{Proceedings of AAAI 2021}, pages 12574--12582. {AAAI} Press.

\bibitem[{Clark et~al.(1999)Clark, Thompson, and Porter}]{clark1999knowledge}
Peter Clark, John Thompson, and Bruce Porter. 1999.
\newblock \href
  {https://www.aaai.org/Papers/Symposia/Fall/1999/FS-99-02/FS99-02-009.pdf} {A
  knowledge-based approach to question-answering}.
\newblock In \emph{Proceedings of AAAI 1999}, pages 43--51.

\bibitem[{Cohen et~al.(2020)Cohen, Rosenman, and
  Goldberg}]{DBLP:journals/corr/abs-2010-04829}
Amir D.~N. Cohen, Shachar Rosenman, and Yoav Goldberg. 2020.
\newblock \href {https://arxiv.org/abs/2010.04829} {Relation extraction as
  two-way span-prediction}.
\newblock \emph{CoRR}, abs/2010.04829.

\bibitem[{Cui et~al.(2017)Cui, Xiao, Wang, Song, Hwang, and
  Wang}]{DBLP:journals/pvldb/CuiXWSHW17}
Wanyun Cui, Yanghua Xiao, Haixun Wang, Yangqiu Song, Seung{-}won Hwang, and Wei
  Wang. 2017.
\newblock \href {http://www.vldb.org/pvldb/vol10/p565-cui.pdf} {{KBQA:}
  learning question answering over {QA} corpora and knowledge bases}.
\newblock \emph{Proceedings of VLDB 2017}, 10(5):565--576.

\bibitem[{Devlin et~al.(2019)Devlin, Chang, Lee, and
  Toutanova}]{DBLP:conf/naacl/DevlinCLT19}
Jacob Devlin, Ming{-}Wei Chang, Kenton Lee, and Kristina Toutanova. 2019.
\newblock \href {https://www.aclweb.org/anthology/N19-1423} {{BERT:}
  pre-training of deep bidirectional transformers for language understanding}.
\newblock In \emph{Proceedings of NAACL 2019}, pages 4171--4186.

\bibitem[{Du and Cardie(2020)}]{DBLP:conf/emnlp/DuC20}
Xinya Du and Claire Cardie. 2020.
\newblock \href {https://doi.org/10.18653/v1/2020.emnlp-main.49} {Event
  extraction by answering (almost) natural questions}.
\newblock In \emph{Proceedings of EMNLP 2020}, pages 671--683.

\bibitem[{Elazar et~al.(2021)Elazar, Zhang, Goldberg, and
  Roth}]{DBLP:journals/corr/abs-2104-08161}
Yanai Elazar, Hongming Zhang, Yoav Goldberg, and Dan Roth. 2021.
\newblock \href {https://doi.org/10.18653/v1/2021.emnlp-main.819} {Back to
  square one: Artifact detection, training and commonsense disentanglement in
  the winograd schema}.
\newblock In \emph{Proceedings of EMNLP 2021}, pages 10486--10500. Association
  for Computational Linguistics.

\bibitem[{Gao et~al.(2021)Gao, Yao, and Chen}]{DBLP:conf/emnlp/GaoYC21}
Tianyu Gao, Xingcheng Yao, and Danqi Chen. 2021.
\newblock \href {https://aclanthology.org/2021.emnlp-main.552} {Simcse: Simple
  contrastive learning of sentence embeddings}.
\newblock In \emph{Proceedings of EMNLP 2021}, pages 6894--6910. Association
  for Computational Linguistics.

\bibitem[{Johnson et~al.(2017)Johnson, Douze, and
  J{\'{e}}gou}]{DBLP:journals/corr/JohnsonDJ17}
Jeff Johnson, Matthijs Douze, and Herv{\'{e}} J{\'{e}}gou. 2017.
\newblock \href {http://arxiv.org/abs/1702.08734} {Billion-scale similarity
  search with gpus}.
\newblock \emph{CoRR}, abs/1702.08734.

\bibitem[{Katz and Fodor(1963)}]{Katz1963-KATTSO-3}
Jerrold Katz and Jerry Fodor. 1963.
\newblock \href {https://doi.org/10.2307/411200} {The structure of a semantic
  theory}.
\newblock \emph{Language}, 39:170--210.

\bibitem[{Kejriwal and Shen(2020)}]{DBLP:journals/corr/abs-2011-09159}
Mayank Kejriwal and Ke~Shen. 2020.
\newblock \href {https://arxiv.org/abs/2011.09159} {Do fine-tuned commonsense
  language models really generalize?}
\newblock \emph{CoRR}, abs/2011.09159.

\bibitem[{Khashabi et~al.(2020)Khashabi, Min, Khot, Sabharwal, Tafjord, Clark,
  and Hajishirzi}]{DBLP:conf/emnlp/KhashabiMKSTCH20}
Daniel Khashabi, Sewon Min, Tushar Khot, Ashish Sabharwal, Oyvind Tafjord,
  Peter Clark, and Hannaneh Hajishirzi. 2020.
\newblock \href {https://doi.org/10.18653/v1/2020.findings-emnlp.171}
  {Unifiedqa: Crossing format boundaries with a single {QA} system}.
\newblock In \emph{Proceedings of EMNLP 2020 Findings}, pages 1896--1907.

\bibitem[{Levesque et~al.(2012)Levesque, Davis, and
  Morgenstern}]{levesque2012winograd}
Hector Levesque, Ernest Davis, and Leora Morgenstern. 2012.
\newblock \href {http://www.aaai.org/ocs/index.php/KR/KR12/paper/view/4492}
  {The winograd schema challenge}.
\newblock In \emph{Proceedings of KR 2012}.

\bibitem[{Lin et~al.(2019)Lin, Chen, Chen, and Ren}]{DBLP:conf/emnlp/LinCCR19}
Bill~Yuchen Lin, Xinyue Chen, Jamin Chen, and Xiang Ren. 2019.
\newblock \href {https://doi.org/10.18653/v1/D19-1282} {Kagnet: Knowledge-aware
  graph networks for commonsense reasoning}.
\newblock In \emph{Proceedings of EMNLP-IJCNLP 2019}, pages 2829--2839.

\bibitem[{Lin et~al.(2020)Lin, Lee, Khanna, and Ren}]{DBLP:conf/emnlp/LinLKR20}
Bill~Yuchen Lin, Seyeon Lee, Rahul Khanna, and Xiang Ren. 2020.
\newblock \href {https://www.aclweb.org/anthology/2020.emnlp-main.557} {Birds
  have four legs?! numersense: Probing numerical commonsense knowledge of
  pre-trained language models}.
\newblock In \emph{Proceedings of EMNLP 2020}, pages 6862--6868.

\bibitem[{Lin(2004)}]{lin2004rouge}
Chin-Yew Lin. 2004.
\newblock \href {https://www.aclweb.org/anthology/W04-1013} {Rouge: A package
  for automatic evaluation of summaries}.
\newblock In \emph{Text summarization branches out}, pages 74--81.

\bibitem[{Liu and Singh(2004)}]{liu2004conceptnet}
Hugo Liu and Push Singh. 2004.
\newblock \href {https://doi.org/10.1023/B:BTTJ.0000047600.45421.6d}
  {Conceptnet: a practical commonsense reasoning tool-kit}.
\newblock \emph{BT technology journal}, 22(4):211--226.

\bibitem[{Lv et~al.(2020)Lv, Guo, Xu, Tang, Duan, Gong, Shou, Jiang, Cao, and
  Hu}]{DBLP:conf/aaai/LvGXTDGSJCH20}
Shangwen Lv, Daya Guo, Jingjing Xu, Duyu Tang, Nan Duan, Ming Gong, Linjun
  Shou, Daxin Jiang, Guihong Cao, and Songlin Hu. 2020.
\newblock \href {https://aaai.org/ojs/index.php/AAAI/article/view/6364}
  {Graph-based reasoning over heterogeneous external knowledge for commonsense
  question answering}.
\newblock In \emph{Proceedings of AAAI 2020}, pages 8449--8456.

\bibitem[{Mostafazadeh et~al.(2020)Mostafazadeh, Kalyanpur, Moon, Buchanan,
  Berkowitz, Biran, and Chu-Carroll}]{mostafazadeh-etal-2020-glucose}
Nasrin Mostafazadeh, Aditya Kalyanpur, Lori Moon, David Buchanan, Lauren
  Berkowitz, Or~Biran, and Jennifer Chu-Carroll. 2020.
\newblock \href {https://www.aclweb.org/anthology/2020.emnlp-main.370}
  {{GLUCOSE}: {G}enera{L}ized and {CO}ntextualized story explanations}.
\newblock In \emph{Proceedings of EMNLP 2020}, pages 4569--4586.

\bibitem[{Paul and Frank(2020)}]{DBLP:conf/emnlp/PaulF20}
Debjit Paul and Anette Frank. 2020.
\newblock \href {https://doi.org/10.18653/v1/2020.findings-emnlp.267} {Social
  commonsense reasoning with multi-head knowledge attention}.
\newblock In \emph{Proceedings of the EMNLP 2020, Findings}, pages 2969--2980.

\bibitem[{Rahman and Ng(2012)}]{DBLP:conf/emnlp/RahmanN12}
Altaf Rahman and Vincent Ng. 2012.
\newblock \href {https://www.aclweb.org/anthology/D12-1071} {Resolving complex
  cases of definite pronouns: The winograd schema challenge}.
\newblock In \emph{Proceedings of CoNLL 2012}, pages 777--789.

\bibitem[{Roemmele et~al.(2011)Roemmele, Bejan, and
  Gordon}]{DBLP:conf/aaaiss/RoemmeleBG11}
Melissa Roemmele, Cosmin~Adrian Bejan, and Andrew~S. Gordon. 2011.
\newblock \href {https://www.aaai.org/ocs/index.php/SSS/SSS11/paper/view/2418}
  {Choice of plausible alternatives: An evaluation of commonsense causal
  reasoning}.
\newblock In \emph{Proceedings of AAAI 2011 Spring Symposium}, pages 90--95.

\bibitem[{Sakaguchi et~al.(2020)Sakaguchi, Bras, Bhagavatula, and
  Choi}]{DBLP:conf/aaai/SakaguchiBBC20}
Keisuke Sakaguchi, Ronan~Le Bras, Chandra Bhagavatula, and Yejin Choi. 2020.
\newblock \href {https://aaai.org/ojs/index.php/AAAI/article/view/6399}
  {Winogrande: An adversarial winograd schema challenge at scale}.
\newblock In \emph{Proceedings of AAAI 2020}, pages 8732--8740.

\bibitem[{Sanh et~al.(2022)Sanh, Webson, Raffel, Bach, Sutawika, Alyafeai,
  Chaffin, Stiegler, Raja, Dey, Bari, Xu, Thakker, Sharma, Szczechla, Kim,
  Chhablani, Nayak, Datta, Chang, Jiang, Wang, Manica, Shen, Yong, Pandey,
  Bawden, Wang, Neeraj, Rozen, Sharma, Santilli, Fevry, Fries, Teehan, Scao,
  Biderman, Gao, Wolf, and Rush}]{DBLP:journals/corr/abs-2110-08207}
Victor Sanh, Albert Webson, Colin Raffel, Stephen Bach, Lintang Sutawika, Zaid
  Alyafeai, Antoine Chaffin, Arnaud Stiegler, Arun Raja, Manan Dey, M~Saiful
  Bari, Canwen Xu, Urmish Thakker, Shanya~Sharma Sharma, Eliza Szczechla,
  Taewoon Kim, Gunjan Chhablani, Nihal Nayak, Debajyoti Datta, Jonathan Chang,
  Mike Tian-Jian Jiang, Han Wang, Matteo Manica, Sheng Shen, Zheng~Xin Yong,
  Harshit Pandey, Rachel Bawden, Thomas Wang, Trishala Neeraj, Jos Rozen,
  Abheesht Sharma, Andrea Santilli, Thibault Fevry, Jason~Alan Fries, Ryan
  Teehan, Teven~Le Scao, Stella Biderman, Leo Gao, Thomas Wolf, and Alexander~M
  Rush. 2022.
\newblock \href {https://openreview.net/forum?id=9Vrb9D0WI4} {Multitask
  prompted training enables zero-shot task generalization}.
\newblock In \emph{Proceedings of ICLR 2022}.

\bibitem[{Sap et~al.(2019)Sap, Le~Bras, Allaway, Bhagavatula, Lourie, Rashkin,
  Roof, Smith, and Choi}]{sap2019atomic}
Maarten Sap, Ronan Le~Bras, Emily Allaway, Chandra Bhagavatula, Nicholas
  Lourie, Hannah Rashkin, Brendan Roof, Noah~A Smith, and Yejin Choi. 2019.
\newblock \href {https://doi.org/10.1609/aaai.v33i01.33013027} {{ATOMIC}: an
  atlas of machine commonsense for if-then reasoning}.
\newblock In \emph{Proceedings of AAAI 2019}, pages 3027--3035.

\bibitem[{Sellam et~al.(2020)Sellam, Das, and
  Parikh}]{DBLP:conf/acl/SellamDP20}
Thibault Sellam, Dipanjan Das, and Ankur~P. Parikh. 2020.
\newblock \href {https://doi.org/10.18653/v1/2020.acl-main.704} {{BLEURT:}
  learning robust metrics for text generation}.
\newblock In \emph{Proceedings of ACL 2020}, pages 7881--7892.

\bibitem[{Talmor et~al.(2019)Talmor, Herzig, Lourie, and
  Berant}]{DBLP:conf/naacl/TalmorHLB19}
Alon Talmor, Jonathan Herzig, Nicholas Lourie, and Jonathan Berant. 2019.
\newblock \href {https://www.aclweb.org/anthology/N19-1421} {Commonsenseqa: {A}
  question answering challenge targeting commonsense knowledge}.
\newblock In \emph{Proceedings of NAACL 2019}, pages 4149--4158.

\bibitem[{Usbeck et~al.(2017)Usbeck, Ngomo, Haarmann, Krithara, R{\"{o}}der,
  and Napolitano}]{DBLP:conf/esws/UsbeckNHKRN17}
Ricardo Usbeck, Axel{-}Cyrille~Ngonga Ngomo, Bastian Haarmann, Anastasia
  Krithara, Michael R{\"{o}}der, and Giulio Napolitano. 2017.
\newblock \href {https://doi.org/10.1007/978-3-319-69146-6\_6} {7th open
  challenge on question answering over linked data {(QALD-7)}}.
\newblock In \emph{Proceedings of 4th SemWebEval Challenge at {ESWC} 2017},
  pages 59--69.

\bibitem[{Wolf et~al.(2019)Wolf, Debut, Sanh, Chaumond, Delangue, Moi, Cistac,
  Rault, Louf, Funtowicz, and Brew}]{DBLP:journals/corr/abs-1910-03771}
Thomas Wolf, Lysandre Debut, Victor Sanh, Julien Chaumond, Clement Delangue,
  Anthony Moi, Pierric Cistac, Tim Rault, R{\'{e}}mi Louf, Morgan Funtowicz,
  and Jamie Brew. 2019.
\newblock \href {http://arxiv.org/abs/1910.03771} {Huggingface's transformers:
  State-of-the-art natural language processing}.
\newblock \emph{CoRR}, abs/1910.03771.

\bibitem[{Wu et~al.(2020)Wu, Wang, Yuan, Wu, and Li}]{DBLP:conf/acl/WuWYWL20}
Wei Wu, Fei Wang, Arianna Yuan, Fei Wu, and Jiwei Li. 2020.
\newblock \href {https://doi.org/10.18653/v1/2020.acl-main.622} {Corefqa:
  Coreference resolution as query-based span prediction}.
\newblock In \emph{Proceedings of ACL 2020}, pages 6953--6963.

\bibitem[{Yih et~al.(2015)Yih, Chang, He, and Gao}]{DBLP:conf/acl/YihCHG15}
Wen{-}tau Yih, Ming{-}Wei Chang, Xiaodong He, and Jianfeng Gao. 2015.
\newblock \href {https://doi.org/10.3115/v1/p15-1128} {Semantic parsing via
  staged query graph generation: Question answering with knowledge base}.
\newblock In \emph{Proceedings of ACL 2015}, pages 1321--1331.

\bibitem[{Yih et~al.(2016)Yih, Richardson, Meek, Chang, and
  Suh}]{DBLP:conf/acl/YihRMCS16}
Wen{-}tau Yih, Matthew Richardson, Christopher Meek, Ming{-}Wei Chang, and Jina
  Suh. 2016.
\newblock \href {https://doi.org/10.18653/v1/p16-2033} {The value of semantic
  parse labeling for knowledge base question answering}.
\newblock In \emph{Proceedings of ACL 2016}, pages 201--206.

\bibitem[{Zellers et~al.(2019)Zellers, Bisk, Farhadi, and
  Choi}]{DBLP:conf/cvpr/ZellersBFC19}
Rowan Zellers, Yonatan Bisk, Ali Farhadi, and Yejin Choi. 2019.
\newblock \href
  {https://openaccess.thecvf.com/content_CVPR_2019/html/Zellers_From_Recognition_to_Cognition_Visual_Commonsense_Reasoning_CVPR_2019_paper.html}
  {From recognition to cognition: Visual commonsense reasoning}.
\newblock In \emph{Proceedings of CVPR 2019}, pages 6720--6731.

\bibitem[{Zhang et~al.(2020{\natexlab{a}})Zhang, Liu, Pan, Song, and
  Leung}]{zhang2019aser}
Hongming Zhang, Xin Liu, Haojie Pan, Yangqiu Song, and Cane Wing-Ki Leung.
  2020{\natexlab{a}}.
\newblock \href {https://doi.org/10.1145/3366423.3380107} {{ASER}: A
  large-scale eventuality knowledge graph}.
\newblock In \emph{Proceedings of WWW 2020}, pages 201--211.

\bibitem[{Zhang et~al.(2020{\natexlab{b}})Zhang, Zhao, and
  Song}]{DBLP:conf/acl/ZhangZS20}
Hongming Zhang, Xinran Zhao, and Yangqiu Song. 2020{\natexlab{b}}.
\newblock \href {https://doi.org/10.18653/v1/2020.acl-main.508} {Winowhy: {A}
  deep diagnosis of essential commonsense knowledge for answering winograd
  schema challenge}.
\newblock In \emph{Proceedings of ACL 2020}, pages 5736--5745.

\bibitem[{Zhang et~al.(2021)Zhang, Zhao, and
  Song}]{DBLP:journals/corr/abs-2009-12721}
Hongming Zhang, Xinran Zhao, and Yangqiu Song. 2021.
\newblock \href {https://aclanthology.org/2021.crac-1.1} {A brief survey and
  comparative study of recent development of pronoun coreference resolution in
  {E}nglish}.
\newblock In \emph{Proceedings of CRAC@EMNLP 2021}, pages 1--11.

\end{thebibliography}

\clearpage

\appendix

\section{Annotation Details}\label{sec:annotation}

\begin{figure}[h]
    \centering
    \includegraphics[width=0.8\linewidth]{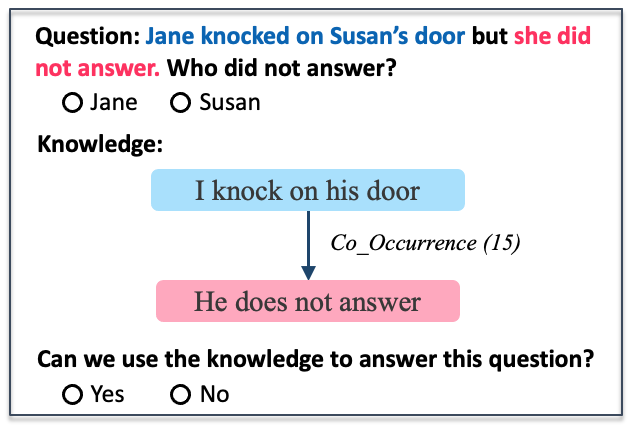}
    \caption{An example of the used survey.}
    \label{fig:survey_demo}
\end{figure}

The annotation goal is to determine whether the supporting graph can help answer the question or not. Thus, for each QA pair, we present the question, candidate answers, and the supporting sub-graph to annotators\footnote{All annotations follow the ethical guidelines.}, and then ask them two questions: (1) What is the correct answer for this question; (2) Whether the provided commonsense knowledge contains all the essential commonsense for answering this question. The purpose of the first question is to assess the annotation quality. A survey example is shown in Figure~\ref{fig:survey_demo}. In beginning of each survey, we also provide detailed instructions and examples to help annotators understand our task. We employ annotators from Amazon Mechanical Turk to provide annotations. To improve the annotation quality, we require the annotators to be English native speaker and to have an overall acceptance rate above 90\%. For each survey, we invite five annotators to provide the annotations and pay them \$0.1.
The average Inter-annotator agreement (Cohen's kappa statistic) for Q1 and Q2 are 0.87 and 0.83, respectively.
The annotation results show that humans could provide consistent annotation about whether the knowledge could be used to answer the questions.








\section{Detailed Experimental Results}\label{sec:detailed_experimental_results}

Detailed experimental results are presented in Table~\ref{tab:Commonsense_Task_Results}.

\begin{table*}[t]
    \small
    \centering
    \begin{tabular}{l||c|c|c|c|c|c|c}
    \toprule
        \multirow{2}{*}{Model} & \multicolumn{7}{c}{Number of Training Instances} \\
        & 5 & 10 & 100 & 500 & 1,000 & 5,000 &  11,678 \\
    \midrule
    Chance Performance & 50.00 (0.00) & 50.00 (0.00) & 50.00 (0.00) & 50.00 (0.00) & 50.00 (0.00) & 50.00 (0.00) & 50.00 (0.00)\\
    \midrule
Vanilla LM & 51.16 (1.92) & 55.88 (2.41) & 56.52 (2.37) & 63.67 (2.19) & 66.76 (1.37) & 70.04 (0.58) & 70.11 (0.28)\\
    \midrule
    KagNet~\cite{DBLP:conf/emnlp/LinCCR19} & 53.29 (2.16) & 55.47 (2.74) & 59.92 (3.05) & 61.97 (1.19) & 65.90 (1.54) & 68.90 (1.21) & 71.50 (1.29)\\
    GBR~\cite{DBLP:conf/aaai/LvGXTDGSJCH20} & 51.77 (1.75) & 56.57 (3.13) & 59.92 (2.34) & 63.36 (1.62) & 68.06 (0.35) & 67.10 (0.17) & 71.34 (0.31)\\
    MHKA~\cite{DBLP:conf/emnlp/PaulF20} & 54.89 (2.34) & 60.47 (1.13) & 61.70 (0.41) & 63.82 (0.78) & 67.85 (0.32) & 69.29 (1.58) & 71.30 (1.14)\\
    G2T~\cite{DBLP:conf/aaai/BianH0021} & \textbf{57.25} (0.21) & \textbf{62.41} (0.97) & \textbf{64.02} (0.99) & \textbf{68.54} (0.47) & \textbf{71.55} (0.75) & \textbf{72.36} (0.56) & \textbf{74.28} (0.21)\\
    
    \midrule
    KagNet-gold& 55.21 (3.21) & 64.36 (0.83) & 68.65 (1.64) & 74.28 (1.31) & 79.05 (0.57) & 80.21 (0.84) & 80.20 (0.21)\\
    GBR-gold & 50.53 (1.62) & 66.34 (1.82) & 69.31 (1.33) & 72.94 (0.35) & 76.24 (0.21) & 80.86 (0.21) & 78.85 (0.13)\\
    MHKA-gold & 58.35 (2.67) & 78.54 (1.32) & 78.55 (0.72) & 79.23 (0.64) & 80.53 (0.50) & 80.52 (0.52) & 81.85 (0.15)\\
    G2T-gold & \textbf{61.39} (2.56) & \textbf{80.85} (1.35) & \textbf{82.18} (0.33) & \textbf{82.51} (0.50) & \textbf{84.32} (0.42) & \textbf{85.81} (0.45) & \textbf{85.48} (0.17)\\
    \bottomrule
    \end{tabular}
    \caption{Demonstration of different models with different training instances. We report the average performance of five different random seeds and standard deviation (in brackets). ``-gold'' indicates that the models are trained and tested with instances with gold knowledge. We cannot directly compare them with the normal setting, but it could serve as the upper-bound for our learning paradigm. Best performing models under both settings are indicated with the \textbf{bold} font.}
    \label{tab:Commonsense_Task_Results}
\end{table*}

\end{document}


\maketitle

\section{Annotation}
Apart from conducting experiments on auto-extracted supporting knowledge graph, we also investigate (1) How many supporting graphs can provide efficient knowledge for the question; (2) How this helpful knowledge improve the model's performance? 
To this end, we annotate whether supporting graphs help answer the questions, then conduct experiments on the questions only with useful supporting graphs again. Those supporting graphs are extracted from three popular commonsense knowledge graphs: ConceptNet, GLUCOSE and ASER.
In this section, we introduce the annotation details and discuss annotation results.

\subsection{Survey Design}

\begin{figure}
    \centering
    \includegraphics[width=0.8\linewidth]{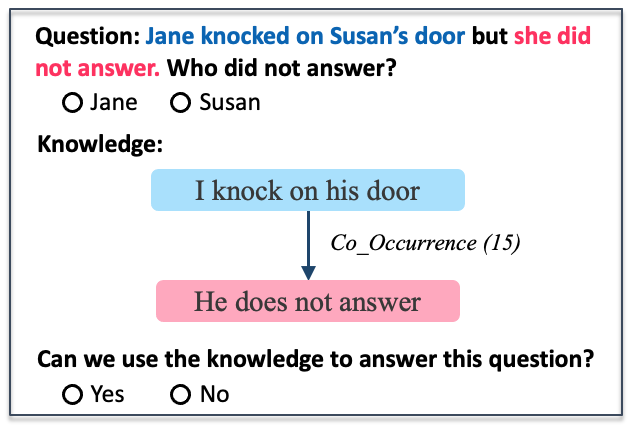}
    \caption{An example of the used survey.}
    \vspace{-0.1in}
    \label{fig:survey_demo}
\end{figure}

The annotation goal is to determine whether the supporting graph can help answer the question or not. Thus, for each QA pair, we present the question, candidate answers, and the supporting sub-graph to annotators, and then ask them two questions: (1) What is the correct answer for this question; (2) Whether the provided commonsense knowledge contains all the essential commonsense for answering this question. The purpose of the first question is to assess the annotation quality. A survey example is shown in Figure~\ref{fig:survey_demo}. In each survey, we also provide detailed instructions and examples to help annotators understand our task.

\subsection{Annotation Details}

We first employ annotators from Amazon Mechanical Turk to provide annotations. To improve the annotation quality, we require the annotators to be English native speaker and to have an overall acceptance rate above 90\%. For each question, we invite five annotators to provide the annotations. 
Considering that our task is relatively challenging for laymen because it does not only involve multiple kinds of commonsense tasks but also involves finding the essential inference path over a graph, we also invite domain experts to annotate the same survey.
Specifically, we invited 30 students to provide the annotation.
After the careful instruction, they can clearly understand the task and requirements.
As a result, they can provide higher quality annotation.
For each QA pair, we randomly assign it to three different students.
To compare these annotation approaches, we show the average accuracy on the first question as well as the inter-annotator agreement on both questions in Table~\ref{tab:annotation_quality}.

The results show that student experts have significantly higher accuracy and agreement than crowd-sourcing workers. This is explainable because unlike an ordinary crowd-sourcing task, our task requires complex thinking and is more challenging for ordinary people.
Considering this, we choose to use the expert annotation as the final label.

\begin{table}[t]
\small
    \centering
    \begin{tabular}{c||c|c}
    \toprule
    &Crowd-sourcing&Experts\\
    \midrule
        Accuracy on Q1 & 0.65 & 0.90 \\
        IAA on Q1 & 0.56 & 0.91\\
        IAA on Q2 & 0.63 & 0.87\\
         \bottomrule
    \end{tabular}
    \caption{Annotation Quality. Q1 and Q2 indicate the original question and the question asking whether the knowledge in the sub-graph is enough or not. IAA is the inter-annotator agreement.}
    \vspace{-0.15in}
    \label{tab:annotation_quality}
\end{table}




\subsection{Annotation Analysis}

\begin{figure*}[tb]
    \centering
	\subfigure[Unhelpful (Not Relevant)]{
		\includegraphics[width=0.35\linewidth]{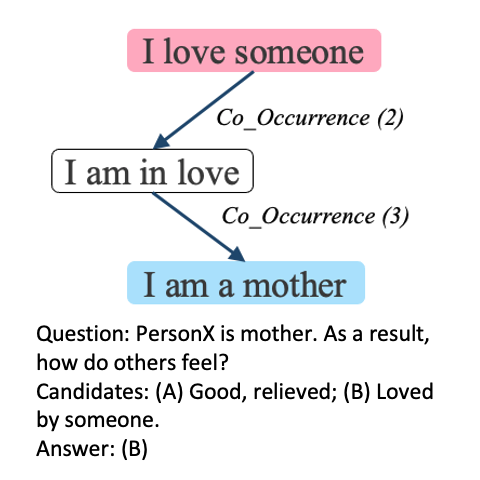}
	}        		
	\subfigure[Unhelpful (Reporting Bias)]{
		\includegraphics[width=0.35\linewidth]{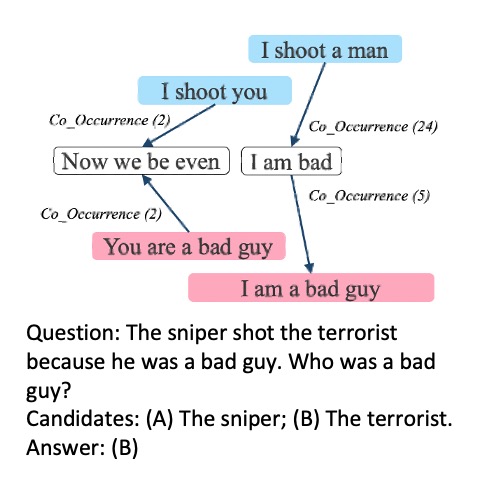}
	}   
		\subfigure[Helpful]{
		\includegraphics[width=0.35\linewidth]{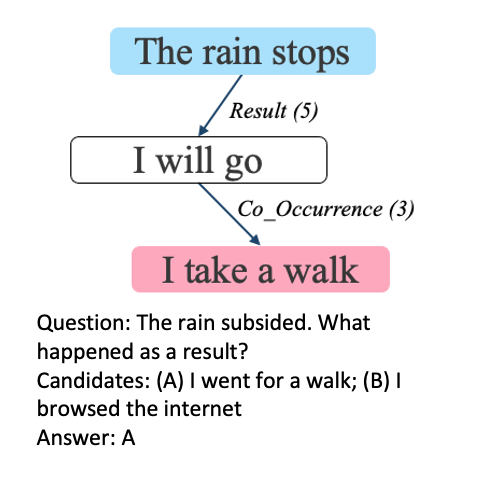}
	}        		
	\subfigure[Helpful]{
		\includegraphics[width=0.35\linewidth]{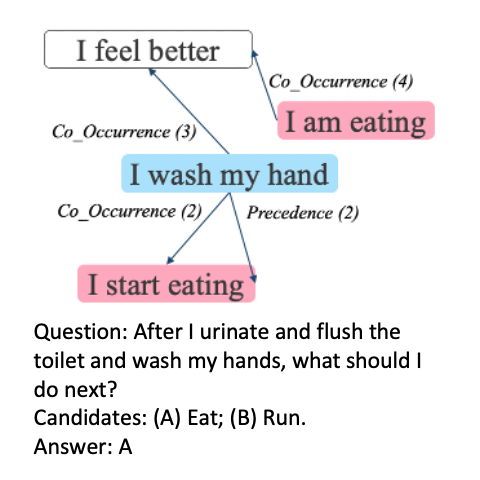}
	}  
    \vspace{-0.1in}
	\caption{Examples of ``Not Helpful'' and ``Helpful'' instances.} 
	\vspace{-0.15in}
	\label{fig:Case_study_appendix}
\end{figure*}

The number of original instances, selected instances (i.e., we can find a relevant sub-graph), and ``helpful'' instances are shown in Table~\ref{tab:CKBQA_statistics_appendix}.
To maximize the dataset quality, for each matching and annotation step, we adopt a very strict standard.
As a result, we sacrifice the recall for precision.
Several ``helpful'' and ``unhelpful'' instances are presented in Figure~\ref{fig:Case_study_appendix}, where (a) and (b) are annotated as ``Not Helpful'' instances and (c) and (d) are annotated as ``Helpful'' ones.
From those examples, we can see that even though all extracted knowledge is relevant with the target questions, some of them are not relevant enough or may contain certain reporting bias.
For example, in Figure~\ref{fig:Case_study_appendix} (a), the question asks how do others feel when someone is a mother, however, the supporting knowledge graph only provides the reaction of the mother. As a result, we could not use such knowledge to answer this question.
In Figure~\ref{fig:Case_study_appendix} (b), the question asks who was the bad guy, the sniper or terrorist. In this case, we can solve this problem because we hold the knowledge that the terrorist is more likely to be the bad guy rather than inferring that knowledge from the ``PersonX shot PersonY'' event.
As a result, if we only look at the two relevant paths in the graph: (1) ``I shoot a man''$\rightarrow$\textit{Co\_Occurrence}$\rightarrow$``I am bad''; (2) ``I shoot you''$\rightarrow$\textit{Co\_Occurrence}$\rightarrow$``Now we be even''$\leftarrow$\textit{Co\_Occurrence}$\leftarrow$``You are a bad guy''. 
We can actually make the wrong prediction because the first path has a higher weight, which is due to the reporting bias of ASER.

As a comparison, we also have some good examples, where the auto-extracted knowledge is indeed very helpful. For example, in Figure~\ref{fig:Case_study_appendix} (c), the question asks about what can be caused when the rain subsided, we are able to find the correct inference path ``The rain stops''$\rightarrow$\textit{Result}$\rightarrow$``I will go''. In Figure~\ref{fig:Case_study_appendix} (d), the question mainly asks what will I do after ``wash my hands,'' and the inference path ``I wash my hand''$\rightarrow$\textit{Precedence}$\rightarrow$``I start eating'' provides the knowledge we needed.

\begin{table*}[t]
    \small
    \centering
    \begin{tabular}{l||c|c|c}
    \toprule
        Task Name  & \# Instances & Avg Sub-graph Size (\# Edges) & \# Helpful Instances\\
    \midrule
   Hard PCR & 2,030 & 3.03 & 185\\
   CommonsenseQA& 530 & 3.30 & 47\\
   COPA & 103 & 2.73 &27\\
   ATOMIC &5,655 & 2.77& 914\\
       \midrule
       Total & 8,318 & 2.87 & 1,173\\
    \bottomrule
    \end{tabular}
    \caption{Detailed \name-ASER statistics.}
    \label{tab:CKBQA_statistics_appendix}
\end{table*}

\subsection{Annotation Statistics}

We report the number of questions that a supported graph can be find, the average size of supporting graph, and the number of helpful instances of \name-ASER in Table~\ref{tab:CKBQA_statistics_appendix}. 
In total, we collect 8,318 instances, and among which Hard PCR and ATOMIC provide the most questions because their original datasets are much larger than others.
According to the annotation, 14.10\% of the supporting knowledge graphs can be used to answer the questions. 
Based on our analysis, annotators hold a very strict standard for whether the supporting knowledge is helpful or not.
For each task, we randomly split the dataset into training, development, and testing set with a standard 8:1:1 splitting.
As a result, we get 6,656 training, 831 development, and 831 testing instances.

\section{Question Transformation Details}
In this section, we demonstrate how we transform questions in selected tasks. We select three datasets for the HardPCR task and their transformation details are shown in Table~\ref{tab:Transformation_for_HardPCR}. For the COPA dataset, there are two question types and we raise different questions for them, the details are in Table~\ref{tab:Transformation_for_COPA}. For the CommonsenseQA task, we use the original question and randomly sample a negative candidate to form a binary choice problem, which you can see in Table~\ref{tab:Transformation_for_CommonsenseQA}. For the ATOMIC dataset, there exist nine different relations in the original dataset thus we formulate questions for each of them, separately. The negative examples are also randomly selected from answers to other questions. Details are in the Table~\ref{tab:Transformation_for_ATOMIC}.

\section{Representative Connectives}
\yt{To Hongming: Extend to other two KBs? We need to renew it?}
As mentioned in the main paper, we convert the knowledge graphs into a paragraph with the help of representative connectives for each discourse relation. The selected connectives for all 15 relations are as follows.
\{``Precedence'': ``before'', ``Succession'': ``after'', ``Synchronous'': ``simultaneously'', ``Reason'': ``because'', ``Result'', ``so'', ``Condition'': ``if'', ``Contrast'': ``but'', ``Concession'': ``although'', ``Conjunction'': ``and'', ``Instantiation'': ``for example'', ``Restatement'': ``in other words'', ``Alternative'': ``or'', ``ChosenAlternative'': ``instead'', ``Exception'': ``except'', ``Co\_Occurrence'': ``and''\}

\section{Implementation Details}
\yt{Renew or delete?}
We implement the experiments with Huggingface and select BERT and RoBERTa as the language models.
As all our models are built on top of the pre-trained languages, so the number of parameters is 110M for BERT-base and RoBERTa-base, and 340M for BERT-large and RoBERTa-large. We use Cross-Entropy as the loss function and Adam as the optimization method. All models are trained for 10,000 steps\footnote{All models converge at 10,000 steps.} and the best-performing checkpoints on the dev set are evaluated.
We tried training with different training data scales to show the learning curves of different settings.
All other hyper-parameters are the same as the base LMs.
We run the experiments on RTX 2080. Due to the small size of datasets, it takes up to 30 minutes for us to train the model for each experiment setting.

\begin{table*}[t]
    \small
    \centering
    \begin{tabular}{l|p{4.0cm}|p{4.0cm}|p{3.0cm}}
    \toprule
     Original Dataset & Original Assertion & Transformed Question & Answer\\
    \midrule
    WSC & Paul tried to call George on the phone, but he wasn't successful. & Paul tried to call George on the phone, but he wasn't successful. Who wasn't successful? & (A) \Blue{Paul}; (B) \Red{George}.\\
    \hline
         DPR & I gave the donkey some food because it was fresh. & I gave the donkey some food because it was fresh. What was fresh? & (A) \Blue{Some food}; (B) \Red{The donkey}.\\
    \hline
        WinoGrande & Jessica wasn't able to stand nearly as long as Maria, since \_ always had strong legs. & Jessica wasn't able to stand nearly as long as Maria, since \_ always had strong legs. Who does \_ refer to? & (A) \Red{Jessica}; (B) \Blue{Maria}.\\
    \bottomrule  
    \end{tabular}
    \caption{Demonstrations of how we transform questions for HardPCR task. Correct answers are in blue.}
    \vspace{-0.1in}
    \label{tab:Transformation_for_HardPCR}
\end{table*}

\begin{table*}[t]
    \small
    \centering
    \begin{tabular}{p{4.0cm}|l|p{4.0cm}|p{2.5cm}}
    \toprule
          Original Assertion & Ask For & Transformed Question & Answer  \\
    \midrule
         I put my plate in the sink. & Cause & I put my plate in the sink. What was the cause of this? & (A) \Blue{I finished eating}; (B) \Red{I skipped dinner}.\\
    \hline
         The rain subsided. & Effect & The rain subsided. What happened as a result? & (A) \Blue{I went for a walk}; (B) \Red{I browsed the internet.}\\
    \bottomrule
    \end{tabular}
    \caption{Demonstrations of how we transform questions for COPA task. Correct answers are in blue.}
    \vspace{-0.1in}
    \label{tab:Transformation_for_COPA}
\end{table*}

\begin{table*}[t]
    \small
    \centering
    \begin{tabular}{p{4.5cm}|p{2.5cm}|p{3.0cm}|p{2.5cm}}
    \toprule
         Original Question & Original Answer & Transformed Question & Sampled Answer  \\
    \midrule
          He picked up his pace to a run, he wanted to do what? & (A) Learn to walk; (B) Frightened; (C) Get away from; (D) Exercise; (E) Go faster. & He picked up his pace to a run, he wanted to do what? & (A) \Red{Learn to walk}; (B) \Blue{Go faster}.\\
    \bottomrule
    \end{tabular}
    \caption{Demonstrations of how we transform questions for CommonsenseQA task. Correct answers are in blue.}
    \vspace{-0.1in}
    \label{tab:Transformation_for_CommonsenseQA}
\end{table*}

\begin{table*}[t]
    \small
    \centering
    \begin{tabular}{p{4.5cm}|p{1.0cm}|p{2.3cm}|p{3.5cm}|p{2.5cm}}
    \toprule
          Original Assertion & Ask For & Original Answer  & Transformed Question & Sampled Answer \\
    \midrule
         PersonX answers PersonY question. & xIntent &  To help person. & PersonX answers PersonY question. Before, what did PersonX want? & (A) \Blue{To help person}; (B) \Red{To further their education}. \\
    \hline
           PersonX becomes a chef. & xNeed & Go to culinary school. & PersonX becomes a chef. Before, what did PersonX need? & (A) \Red{To be near a vending machine}; (B) \Blue{Go to culinary school}.\\
    \hline
           PersonX becomes afraid. & xAttr & Scared. & PersonX becomes afraid. What is PersonX seen as? & (A) \Blue{Scared}; (B) \Red{Motivating}.\\
    \hline
           PersonX finds a new job.  & xReact & Satisfied. &PersonX finds a new job. As a result, how does PersonX feel? & (A) \Blue{Satisfied}; (B) \Red{Nervous}.\\
    \hline
           PersonX gets hungry. & xWant & Find and grab food. & PersonX gets hungry. As a result, what does PersonX want?& (A) \Blue{Find out the results}; (B) \Red{Find and grab food}.\\
 \hline
           PersonX gets ready for bed. & xEffect & Goes to bed. & PersonX gets ready for bed. What effect does the event have on PersonX? & (A) \Red{Uses thing}; (B) \Blue{Goes to bed}. \\
\hline
           PersonX gives PersonY confidence. & oReact & Happy. & PersonX gives PersonY confidence. As a result, how do others feel? & (A) \Blue{Happy}; (B) \Red{Sad}. \\
\hline
           PersonX goes deaf. & oWant & To get a hearing aid for him. & PersonX goes deaf. As a result, what do others want? & (A) \Blue{To get a hearing aid for him}; (B) \Red{To go home}.\\
\hline
           PersonX teaches PersonY so much. & oEffect & Respect their knowledge. & PersonX teaches PersonY so much. What effect does the event have on others? & (A) \Red{They lose sleep}; (B) \Blue{Respect their knowledge}.\\
    \bottomrule
    \end{tabular}
    \caption{Demonstrations of how we transform questions for ATOMIC task. Correct answers are in blue.}
    \vspace{-0.1in}
    \label{tab:Transformation_for_ATOMIC}
\end{table*}